\documentclass{article}

\usepackage{arxiv}

\usepackage[utf8]{inputenc} 
\usepackage[T1]{fontenc}    
\usepackage{hyperref}       
\usepackage{url}            
\usepackage{booktabs}       
\usepackage{amsfonts}       
\usepackage{nicefrac}       
\usepackage{microtype}      
\usepackage{lipsum}
\usepackage{amsfonts}
\usepackage{amsthm}
\usepackage{amsmath}
\usepackage{tikz}
\usepackage{caption}
\newcommand\numberthis{\addtocounter{equation}{1}\tag{\theequation}}

\title{Using the Naive Bayes as a discriminative classifier}

\author{
Elie Azeraf\thanks{Elie Azeraf is also a member of SAMOVAR, Telecom SudParis, Institut Polytechnique de Paris} \\
  Watson Department \\
  IBM GSB France \\
  \texttt{elie.azeraf@ibm.com} \\
   \And
Emmanuel Monfrini \\
SAMOVAR, Telecom SudParis \\
Institut Polytechnique de Paris \\
\And
Wojciech Pieczynski \\
SAMOVAR, Telecom SudParis \\
Institut Polytechnique de Paris \\
}

\begin{document}
\maketitle

\begin{abstract}
For classification tasks, probabilistic models can be categorized into two disjoint classes: generative or discriminative. It depends on the posterior probability computation of the label $x$ given the observation $y$, $p(x | y)$. On the one hand, generative classifiers, like the Naive Bayes or the Hidden Markov Model (HMM), need the computation of the joint probability p(x,y), before using the Bayes rule to compute $p(x | y)$. On the other hand, discriminative classifiers compute $p(x | y)$ directly, regardless of the observations' law. They are intensively used nowadays, with models as Logistic Regression, Conditional Random Fields (CRF), and Artificial Neural Networks. However, the recent Entropic Forward-Backward algorithm shows that the HMM, considered as a generative model, can also match the discriminative one's definition. This example leads to question if it is the case for other generative models. In this paper, we show that the Naive Bayes classifier can also match the discriminative classifier definition, so it can be used in either a generative or a discriminative way. Moreover, this observation also discusses the notion of Generative-Discriminative pairs, linking, for example, Naive Bayes and Logistic Regression, or HMM and CRF. Related to this point, we show that the Logistic Regression can be viewed as a particular case of the Naive Bayes used in a discriminative way. 
\end{abstract}


\section{Introduction}

Given an input data y, also called observation, and a discrete finite output space $\Lambda_X = \{ \lambda_1, ..., \lambda_N \}$, a classification task consists in computing the different label probabilities $p(x = \lambda_i  | y)$. They can be used in many areas as Natural Language Processing with Text Classification \cite{kowsari2019text} or Sentiment Analysis \cite{maas-etal-2011-learning}, or Computer Vision with Image Classification \cite{deng2009imagenet} or Object Detection \cite{lin2014microsoft}, among many others. 
These model classifiers are usually categorized into two disjoint classes: generative or discriminative. From \cite{ng2002discriminative}, a generative classifier learns the joint probability $p(x,y)$, then uses the Bayes rule to compute the posterior probability $p(x | y)$:
\begin{align}
\forall \lambda_i \in \Lambda_X, p(x = \lambda_i | y) = \frac{p(x = \lambda_i, y)}{\sum_{j \in \Lambda_X} p(x = \lambda_j, y)}
\end{align}

To train a generative model in a supervised context, one has to maximize the joint likelihood $p(x,y)$ of the training data. For example, in the case where y is a discrete variable, it consists in counting the different patterns’ occurrence. Among the most popular generative models, we can cite the Naive Bayes \cite{lewis1998naive, mccallum1998comparison, rish2001empirical, Webb2010}, the Hidden Markov Model (HMM) \cite{cappe2009inference, rabiner1986introduction, stratonovich1965conditional}, or the Gaussian Mixture Model \cite{rasmussen2000infinite, Reynolds2009}.

On the other hand, a discriminative classifier computes $p(x | y)$ directly, which means that p(x,y) does not have to be known. We can cite the Logistic Regression \cite{friedman2001elements, hosmer2013applied, menard2002applied, peng2002introduction, wright1995logistic}, the Maximum Entropy Markov Model (MEMM) \cite{mccallum2000maximum}, the Conditional Random Fields (CRF) \cite{lafferty2001conditional, sutton2006introduction}, or even the Artificial Neural Networks (ANN) \cite{goodfellow2016deep, lecun2015deep}, among the most popular ones. They are usually trained by fitting the posterior distribution $p(x | y)$, or minimizing a loss function, thanks to optimization algorithms like the gradient descent \cite{ruder2016overview}. 

Many papers use these definitions \cite{bishop2006pattern, bouchard2004tradeoff, lasserre2006principled, minka2005discriminative, roth2018hybrid, sutton2006introduction, ulusoy2005generative, yakhnenko2005discriminatively} and compare both approaches, with a general preference for the second category where the task does not have unlabeled data. Indeed, generative classifiers’ main criticism concerns their learning strategy, imposing to learn the joint probability, and therefore the observation's distribution. This learning method constrains the observation's features during classification tasks. Indeed, it is considered impossible to handle arbitrary features with a generative model \cite{jurafsky2000speech, mccallum2000maximum, sutton2006introduction}, except with a restrictive independence condition. The next section will illustrate this point.

\begin{figure}
\begin{center}
\begin{tikzpicture}
[font=\small, inner sep=0pt, hidden/.style = {circle,draw = blue!15, fill = blue!10, thick,minimum size = 0.75cm, rounded corners}, visible/.style = {circle,draw=black!35,fill=black!30,thick,minimum size=0.75cm, rounded corners}, scale = 0.75]
\node at (-4.5,2.25) (x1) [visible] {$x_1$};
\node at (-4.5,0) (h1) [hidden]  {$y_1$};
\draw [->, >=stealth] (x1) to (h1);
                    
\node at (-2.25,2.25) (x2) [visible] {$x_2$};
\node at (-2.25,0) (h2) [hidden]  {$y_2$};
\draw [->, >=stealth] (x2) to (h2);
                    
\node at (0,2.25) (x3) [visible] {$x_3$};
\node at (0,0) (h3) [hidden]  {$y_3$};
\draw [->, >=stealth] (x3) to (h3);
                    
\node at (2.25, 2.25) (x4) [visible] {$x_4$};
\node at (2.25,0) (h4) [hidden]  {$y_4$};
\draw [->, >=stealth] (x4) to (h4);
                    
\draw [->, >=stealth] (x1) to (x2);
\draw [->, >=stealth] (x2) to (x3);
\draw [->, >=stealth] (x3) to (x4);
\end{tikzpicture}
\captionof{figure}{Probabilistic oriented graph of the HMM}
\label{fig_hmm}
\end{center}
\end{figure}
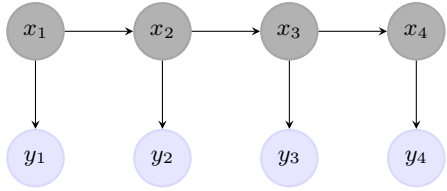

We focus on the HMM, represented in figure \ref{fig_hmm}, with an observed process $y_{1:T} = (y_1,...,y_T)$, and a hidden one $x_{1:T} = (x_1,...,x_T)$. This model is considered as a generative probabilistic one. Indeed, if one wants to compute $p(x_t = \lambda_i | y_{1:T})$, he can apply the Forward-Backward algorithm \cite{rabiner1986introduction, rabiner1989tutorial} described in the appendices. In agreement with the generative model definition, this algorithm firstly computes the joint probability $p(x_t = \lambda_i, y_{1:T} )$ before computing the posterior one. Moreover, all the parameters are learned with maximum likelihood. However, a recent work about HMM presents the Entropic Forward-Backward algorithm \cite{azeraf2020hidden}, also described in the appendices. It allows computing the posterior distribution $p(x_t = \lambda_i | y_{1:T})$ directly, with using neither the joint distribution, nor the observations’ one. It also allows training the HMM with optimization algorithms as the gradient descent. Therefore, in this case, the HMM matches the discriminative classifier’s definition.

This HMM example leads to question this categorization of probabilistic classifiers, which is the aim of the paper. We present two contributions:
\begin{enumerate}
    \item We show that the Naive Bayes, a popular generative model, can also match the definition of a discriminative one. It is another example showing that the traditional definitions of both discriminative and generative models do not necessarily lead to disjoint categories.
    \item We discuss the notion of Generative-Discriminative pairs \cite{ng2002discriminative, sutton2006introduction}, linking a generative model with its discriminative counterpart, as for example, the Naive Bayes - Logistic Regression pair, or the HMM - CRF one. Considering the first example, we show that the Logistic Regression is a particular case of the Naive Bayes used in a discriminative way. 
\end{enumerate}

This paper is organized as follows. In the next section, we recall the Naive Bayes model's law with its oriented probabilistic graph, and we present the generative way, usually used, to compute the label probabilities. Then, we show how to compute these probabilities in a discriminative way, which does not use the observation's law. In the third section, after some recalls about the Logistic Regression model, we show that this latter can be viewed as a particular case of the Naive Bayes used in a discriminative way. Conclusion and perspectives lie at the end of the paper.

\section{The Naive Bayes classifier as a discriminative model}

\subsection{Naive Bayes classifier}

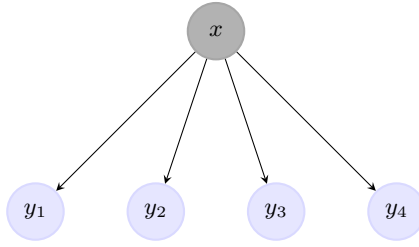
\begin{figure}
\begin{center}
\begin{tikzpicture}
[font=\small, inner sep=0pt, hidden/.style = {circle,draw = blue!15, fill = blue!10, thick,minimum size = 0.75cm, rounded corners}, visible/.style = {circle,draw=black!35,fill=black!30,thick,minimum size=0.75cm, rounded corners}, scale = 0.8]
\node at (0,0) (x) [visible] {$x$};

\node at (-3,-3) (y1) [hidden]  {$y_1$};
\draw [->, >=stealth] (x) to (y1);
                    
\node at (-1,-3) (y2) [hidden]  {$y_2$};
\draw [->, >=stealth] (x) to (y2);
                    
\node at (1,-3) (y3) [hidden]  {$y_3$};
\draw [->, >=stealth] (x) to (y3);

\node at (3,-3) (y4) [hidden]  {$y_4$};
\draw [->, >=stealth] (x) to (y4);
\end{tikzpicture}
\end{center}
\captionof{figure}{Probabilistic graphical representation of the Naive Bayes}
\label{fig_nb}
\end{figure}

The Naive Bayes is a probabilistic graphical model \cite{koller2009probabilistic} considering the observations $y_{1:T} = (y_1, ..., y_T)$, with $y_t$ taking its value in $\Omega_{Y_t}$, and a hidden variable x taking its value in $\Lambda_X$. It models the joint probability of $(x, y_{1:T})$ with the following law:
\begin{align}
p(x,y_{1:T}) = p(x) p(y_1 |x) p(y_2 | x) ... p(y_{T-1} | x) p(y_T | x) = p(x) \prod_{t = 1}^T p(y_t | x)
\end{align}
The oriented probabilistic graph is given in figure \ref{fig_nb}. 

In the generative way, as usually used, the Naive Bayes classifier computes, for each $\lambda_i \in \Lambda_X, p(x=\lambda_i | y_{1:T})$ as follows:
\begin{align*}
p(x = \lambda_i | y_{1:T}) &= \frac{p(x=\lambda_i, y_{1:T}) }{ \sum_{\lambda_j \in \Lambda_X} p(x = \lambda_j, y_{1:T})} \\
&= \frac{ p(x = \lambda_i) \prod_{t=1}^{T}{ p(y_t |  x = \lambda_i)} }{\sum_{\lambda_j\in\Lambda_X}{ p(x = \lambda_j) \prod_{t=1}^{T}{ p(y_t | x = \lambda_j)}}} \\
&= \frac{ \pi(i) \prod_{t=1}^{T}{b_i^{(t)}(y_t)} }{ \sum_{ \lambda_j \in \Lambda_X }{ \pi(j) \prod_{t=1}^{T}{b_j^{(t)}(y_t)}}} \numberthis
\label{eq_gen_nb}
\end{align*}
using the notations:
\begin{itemize}
    \item $\pi(i) = p(x = \lambda_i)$;
    \item $b_i^{(t)}(y) = p(y_t | x_t = \lambda_i)$.
\end{itemize}

With (\ref{eq_gen_nb}), the Naive Bayes effectively matches the generative classifier definition, as it first computes the joint probability $p(x = \lambda_i, y_{1:T})$, and then the posterior $p(x = \lambda_i | y_{1:T})$.

Used in this way, it is facing difficulties to consider arbitrary features. Indeed, let us consider, for example, $\Omega_{Y_t} = \Omega_Y^2$, for each t, with $\Omega_Y = \{ \omega_1, ..., \omega_M\}$. Thus, each observation has two features: $y_t = (\omega_i, \omega_j)$. With this restoration method, the different parameters are learned by maximum likelihood estimation, which consists in counting the different patterns:
\begin{align*}
\pi(i) = \frac{f(i)}{L}, \;\;\;\;\;\;\;\;\;\;\;\;\;\;\; 
b_i^{(t)}(y_t) = \frac{f_i^{(t)}(j,k)}{f(i)},
\end{align*}
with L the number of training samples, $f\left(i\right)$ the number of times $x=\lambda_i$ in this training sample, and $f_i^{\left(t\right)}\left(j,k\right)$ the number of times $y_t=\left(\omega_j,\omega_k\right)$ when $x=\lambda_i$. Therefore, an observation $y_t$ has its value $b_i^{\left(t\right)}\left(y_t\right)$ different from 0 if and only if there is an observation in the training set having the same features at position t. If this estimation method is possible with a small number of features, it quickly becomes intractable when the number of features increases, making mandatory to suppose them independent. It is especially the case for Natural Language Processing tasks, where features can be suffixes of any length, prefixes, some word's characteristics, or large numerical vector issued from an embedding method \cite{almeida2019word}.

\subsection{Computing directly $p\left(x = \lambda_i\middle | y_{1:T}\right)$ with the Naive Bayes classifier}

In this section, we show that one can use the Naive Bayes classifier by computing $p\left(x=\lambda_i\middle|\ y_{1:T}\right)$ directly, without previous computation of $p\left(x=\lambda_i,y_{1:T}\right)$. To simplify notations, we set, for each $\lambda_i\in\Lambda_X$:
\begin{align*}
L_{y_t}^{\left(t\right)}\left(i\right)=p\left(x=\lambda_i\middle|\ y_t\right).
\end{align*}

We can state the following result:
\paragraph{Proposition 1.} Let $p\left(x=\lambda_i\middle|\ y_{1:T}\right)$ be a Naive Bayes distribution. Then for each $\lambda_i\in\mathrm{\Lambda}_X$:
\begin{align}
p\left(x=\lambda_i\middle| y_{1:T}\right)=\frac{\pi\left(i\right)^{1-T}\prod_{t=1}^{T}{L_{y_t}^{\left(t\right)}\left(i\right)}}{\sum_{j=1}^{N}{\pi\left(j\right)^{1-T}\prod_{t=1}^{T}{L_{y_t}^{\left(t\right)}\left(j\right)}}}
\label{eq_nb_dis}
\end{align}

\paragraph{Proof.} For each $\lambda_i\in\mathrm{\Lambda}_X$, we set:
\begin{align}
\delta\left(i\right)=\pi\left(i\right)^{1-T}\prod_{t=1}^{T}{L_{y_t}^{\left(t\right)}\left(i\right)}
\end{align}
This function is linked with the joint probability by:
\begin{align}
p\left(x=\lambda_i,y_{1:T}\right)=p\left(y_1\right)p\left(y_2\right)\ldots p\left(y_{T-1}\right)p\left(y_T\right)\delta\left(i\right)
\end{align}
Indeed,
\begin{align*}
p\left(x=\lambda_i,y_{1:T}\right) &= \pi\left(i\right)\prod_{t=1}^{T}{b_i^{\left(t\right)}\left(y_t\right)}
&= \pi\left(i\right)\prod_{t=1}^{T}\frac{p\left(y_t,x=\lambda_i\right)}{p\left(x=\lambda_i\right)}
&= \pi\left(i\right)^{1-T}\prod_{t=1}^{T}{p\left(y_t\right)L_{y_t}^{\left(t\right)}\left(i\right)}
&=\delta\left(i\right)\prod_{t=1}^{T}p\left(y_t\right)
\end{align*}
Therefore 
\begin{align}
p\left(x=\lambda_i\middle| y_{1:T}\right)=\frac{p\left(x=\lambda_i,y_{1:T}\right)}{\sum_{j=1}^{N}p\left(x=\lambda_j,y_{1:T}\right)}=\frac{\delta\left(i\right)}{\sum_{j=1}^{N}\delta\left(j\right)},
\end{align}
which ends the proof.

Finally, given (\ref{eq_nb_dis}), the Naive Bayes model also matches the definition of a discriminative classifier: it allows computing $p\left(x\middle|\ y_{1:T}\right)$ directly and does use neither the observations' law nor the joint one. Therefore, through this example, we see that in some situations, “discriminative” and “generative” definitions do not describe the deep nature of a model, but rather the way it is used. 
When differentiable functions model $L_{y_t}^{\left(t\right)}$, one can train the Naive Bayes used in a discriminative way with (\ref{eq_nb_dis}) using gradient descent algorithms. In this case, the training process consists in computing the gradient of a given loss function by applying the backpropagation \cite{lecun1989backpropagation, lecun1990handwritten} algorithm, and updating the model's parameters. In this way, the Naive Bayes can be applied for any classification tasks and consider arbitrary observation features. For example, a homogeneous Naive Bayes used in a discriminative way, modelling $L_{y_t}^{\left(t\right)}$ with the same function independent of t, can be applied to classify text of various lengths. First of all, one has to use an embedding method like Flair \cite{akbik2018contextual}, FastText \cite{bojanowski2017enriching}, or BERT \cite{devlin2018bert}, to transform word into numerical vector. Then, he has to select a function to model $L_{y_t}^{\left(t\right)}$, for example, a feedforward neural network.

\section{The Logistic Regression: a particular case of the NaIve Bayes used in a discriminative way}

\subsection{Logistic Regression classifier}

Let us consider multinomial Logistic Regression, which is the more general one. For each $t, \Omega_{Y_t}=\mathbb{R}$, Logistic Regression computes $p\left(x=\lambda_i\middle|\ y_{1:T}\right)$ with
\begin{align}
p\left(x=\lambda_i\middle| y_{1:T}\right)=\frac{\exp{\left(W_i\cdot y_{1:T}+b_i\right)}}{\sum_{j=1}^{N}\exp{\left(W_j\cdot y_{1:T}+b_j\right)}}
\label{eq_lr}
\end{align}
where for each $i\in\{1,\ldots,N\},\ W_i\in\mathbb{R}^T, b_i\in\mathbb{R}$, and $\cdot$ denotes the matrix product.

\begin{figure}
\begin{center}
\begin{tikzpicture}
[font=\small, inner sep=0pt, hidden/.style = {circle,draw = blue!15, fill = blue!10, thick,minimum size = 0.75cm, rounded corners}, visible/.style = {circle,draw=black!35,fill=black!30,thick,minimum size=0.75cm, rounded corners}, scale = 0.8]
\node at (0,0) (x) [visible] {$x$};

\node at (-3,-3) (y1) [hidden]  {$y_1$};
\draw [-, >=stealth] (x) to (y1);
                    
\node at (-1,-3) (y2) [hidden]  {$y_2$};
\draw [-, >=stealth] (x) to (y2);
                    
\node at (1,-3) (y3) [hidden]  {$y_3$};
\draw [-, >=stealth] (x) to (y3);

\node at (3,-3) (y4) [hidden]  {$y_4$};
\draw [-, >=stealth] (x) to (y4);
\end{tikzpicture}
\end{center}
\captionof{figure}{Probabilistic graphical representation of the Logistic Regression}
\label{fig_lr}
\end{figure}
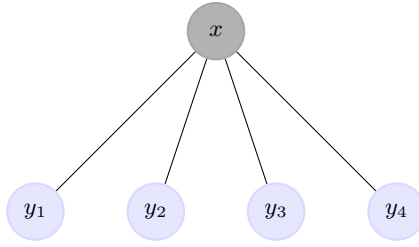

Figure \ref{fig_lr} represents the oriented graph of this classifier.

The Logistic Regression is a probabilistic graphical model used in numerous applications. One can train it with a gradient descent algorithm. It is considered as a discriminative model as it computes directly $p\left(x\middle|\ y_{1:T}\right)$. With this model, we notice that each probability computed is strictly greater than 0.

\subsection{Logistic Regression and Naive Bayes}

Logistic Regression is frequently linked with the Naive Bayes with the notion of Generative – Discriminative pairs \cite{ng2002discriminative, sutton2006introduction}, assuming it is the discriminative counterpart of the Naive Bayes. However, we are going to show that the Logistic Regression is a particular case of the Naive Bayes.

We can state:
\paragraph{Proposition 2.} Let us consider a Naive Bayes classifier used in a discriminative way with (\ref{eq_nb_dis}). If, for each $i\ \in\{1,\ \ldots,\ N\},\ t\in{1,\ \ldots,\ T}$:
\begin{align}
L_{y_t}^{\left(t\right)}\left(i\right)=\frac{exp{\left(a_i^{\left(t\right)}y_t+c_i^{\left(t\right)}\right)}}{\sum_{k=1}^{N}exp{\left(a_k^{\left(t\right)}y_t+c_k^{\left(t\right)}\right)}}
\end{align}
with $a_i^{\left(t\right)}\in\mathbb{R}$ and $c_i^{\left(t\right)}\in\mathbb{R}$, then this Naive Bayes classifier is a Logistic Regression. The reciprocal is also true.

\paragraph{Proof} We start from (\ref{eq_nb_dis}), for each $\lambda_i \in \Lambda_X$:
\begin{align*}
p(x = \lambda_i | y_{1:T}) &= \frac{ \pi(i)^{1-T} \prod_{t=1}^{T} {L_{y_t}^{\left(t\right)}\left(i\right)}}{\sum_{j=1}^{N}{\pi\left(j\right)^{1-T}\prod_{t=1}^{T}{L_{y_t}^{\left(t\right)}\left(j\right)}}} \\
&= \frac{\pi\left(i\right)^{1-T}\prod_{t=1}^{T}\frac{exp{\left(a_i^{\left(t\right)}y_t+c_i^{\left(t\right)}\right)}}{\sum_{k=1}^{N}exp{\left(a_k^{\left(t\right)}y_t+c_k^{\left(t\right)}\right)}}}{\sum_{j=1}^{N}{\pi\left(j\right)^{1-T}\prod_{t=1}^{T}\frac{exp{\left(a_j^{\left(t\right)}y_t+c_j^{\left(t\right)}\right)}}{\sum_{k=1}^{N}exp{\left(a_k^{\left(t\right)}y_t+c_k^{\left(t\right)}\right)}}}} \\
&= \frac{\pi\left(i\right)^{1-T}\prod_{t=1}^{T}exp{\left(a_i^{\left(t\right)}y_t+c_i^{\left(t\right)}\right)}}{\sum_{j=1}^{N}{\pi\left(j\right)^{1-T}\prod_{t=1}^{T}exp{\left(a_j^{\left(t\right)}y_t+c_j^{\left(t\right)}\right)}}} \\
&= \frac{exp{\left(\left(1-T\right)log{\left(\pi\left(i\right)\right)}+\sum_{t=1}^{T}{a_i^{\left(t\right)}y_t}+c_i^{\left(t\right)}\right)}}{\sum_{j=1}^{N}exp{\left(\left(1-T\right)log{\left(\pi\left(j\right)\right)}+\sum_{t=1}^{T}{a_j^{\left(t\right)}y_t}+c_i^{\left(t\right)}\right)}}
\end{align*}

We set:
\begin{itemize}
    \item $b_i=\left(1-T\right)log{\left(\pi\left(i\right)\right)}+\sum_{t=1}^{T}c_i^{\left(t\right)}$;
    \item $W_i=\left[a_i^{\left(1\right)},a_i^{\left(2\right)},\ldots,a_i^{\left(T\right)}\right]$.
\end{itemize}
Therefore, we can effectively verify that $b_i\in\mathbb{R}$ and $W_i\in\mathbb{R}^T$.

Then:
\begin{align}
p\left(x=\lambda_i\middle| y_{1:T}\right)=\frac{exp{\left(W_i\cdot y_{1:T}+b_i\right)}}{\sum_{j=1}^{N}exp{\left(W_j\cdot y_{1:T}+b_j\right)\ \ \ \ }}
\end{align}
which is the equation (\ref{eq_lr}) defining a Logistic Regression. Proving the reciprocal is straightforward. Indeed, given the parameters of a logistic regression $W_i$ and $b_i$, it is always possible to define the parameters $a_i^{(t)}, \pi(i)$, and $c_i^{(t)}$ related to a Naive Bayes.

Proposition 2 shows that one can view the Logistic Regression as a particular case of the Naive Bayes used in a discriminative way.

\section{Conclusion}

This paper presents how to use the Naive Bayes as a discriminative model. A practical consequence is that it can be used in both generative and discriminative ways; the latter allowing arbitrary feature consideration. Moreover, we show that the Logistic Regression, usually presented as the discriminative counterpart of the Naive Bayes, can be seen as a particular case of the latter used in a discriminative way. A general conclusion is that the usual definitions of generative and discriminative models do not lead to disjoint families; and there are models that simultaneously satisfy both of them. An interesting perspective would consist in extending ideas of the paper to other popular generative models and examining whether they can also be used in a discriminative way. 

\bibliographystyle{unsrt}  
\bibliography{references}

\section*{Appendix}

In this appendix, we present both HMM's algorithms to restore, for each $\lambda_i\in\Lambda_X,\ t\in\{1,\ldots,T\},\ y_{1:T}\in{(\Omega}_Y)^T,\  p\left(x_t=\lambda_i\middle|\ y_{1:T}\right)$: the Forward-Backward algorithm, matching the generative definition, and the Entropic Forward-Backward, which matches the discriminative one. We consider a homogeneous HMM with the following notations:
\begin{itemize}
    \item $\pi\left(i\right)=p\left(x_t=\lambda_i\right)$;
    \item $a_i\left(j\right)=p\left(x_{t+1}=\lambda_j\middle|\ x_t=\lambda_i\right)$;
    \item $b_i\left(y_t\right)=p\left(y_t\middle|\ x_t=\lambda_i\right)$;
    \item $L_{y_t}\left(i\right)=p\left(x_t=\lambda_i\middle|\ y_t\right)\ $.
\end{itemize}

\subsection*{A. HMM as a generative classifier: The Forward-Backward algorithm}

The Forward-Backward algorithm consists in computing $p\left(x_t=\lambda_i\middle|\ y_{1:T}\right)$ as follows:
\begin{align*}
p\left(x_t=\lambda_i\middle|\ y_{1:T}\right)=\frac{\alpha_t\left(i\right)\beta_t\left(i\right)}{\sum_{j=1}^{N}{\alpha_t\left(j\right)\beta_t\left(j\right)}}
\end{align*}
with forward probabilities $\alpha$ computed with the following recursion:
\begin{align*}
\alpha_1\left(i\right) &= \pi\left(i\right)b_i\left(y_1\right) \\
\alpha_{t+1}\left(i\right) &= b_i\left(y_{t+1}\right)\sum_{j=1}^{N}{\alpha_t\left(j\right)a_j\left(i\right)}
\end{align*}

And the backward probabilities $\beta$:
\begin{align*}
\beta_T\left(i\right) &= 1 \\ 
\beta_t\left(i\right) &= \sum_{j=1}^{N}{\beta_{t+1}\left(j\right)a_i\left(j\right)b_j\left(y_{t+1}\right)}
\end{align*}

\subsection{B. HMM as a discriminative classifier: The Entropic Forward-Backward algorithm}

The Entropic Forward-Backward algorithm consists in computing, $p\left(x_t=\lambda_i\middle|\ y_{1:T}\right)$ as follows:
\begin{align*}
p\left(x_t=\lambda_i\middle|\ y_{1:T}\right)=\frac{\alpha_t^E\left(i\right)\beta_t^E\left(i\right)}{\sum_{j=1}^{N}{\alpha_t^E\left(j\right)\beta_t^E\left(j\right)}}
\end{align*}
with entropic forward probabilities $\alpha^E$ computed as:
\begin{align*}
\alpha_1^E\left(i\right) &= L_{y_1}\left(i\right) \\
\alpha_{t+1}^E\left(i\right) &= \frac{L_{y_{t+1}}\left(i\right)}{\pi\left(i\right)}\sum_{j=1}^{N}{\alpha_t^E\left(j\right)a_{j\left(i\right)}}
\end{align*}
And the entropic backward $\beta^E$:
\begin{align*}
\beta_T^E\left(i\right) &= 1 \\
\beta_t^E\left(i\right) &= \sum_{j=1}^{N}\frac{L_{y_{t+1}}\left(j\right)}{\pi\left(j\right)}\beta_{t+1}^E(j)\ a_i(j)
\end{align*}

\end{document}